\documentclass[conference]{IEEEtran}

\makeatletter
\newcommand{\linebreakand}{%
  \end{@IEEEauthorhalign}
  \hfill\mbox{}\par
  \mbox{}\hfill\begin{@IEEEauthorhalign}
}
\makeatother

\usepackage[switch]{lineno}

\usepackage{amsmath,amssymb,amsthm,amsfonts,mathrsfs}
\usepackage[usenames,svgnames,x11names,dvipsnames]{xcolor}
\usepackage{xspace} 
\usepackage{soul}
\usepackage{xr-hyper} 
\usepackage{hyperref} 
\sethlcolor{gray}
\usepackage{booktabs,colortbl}
\usepackage{makecell}

\usepackage{tikz}
\usepackage{pgfplots,pgfplotstable}
\usetikzlibrary{
    shapes,
    arrows.meta,
    backgrounds,
    intersections,
    external,
    calc,
}
\usepgfplotslibrary{groupplots, fillbetween,}
\pgfplotsset{compat=1.8}

\newcommand{\paren}[1]{\left(#1\right)}
\newcommand{\set}[1]{\left\{#1\right\}}

\newcommand{\ADDCITATION}[1]{\textcolor{red}{ADDCITATION}}
\newcommand{\single}{CAE\xspace}
\newcommand{\double}{BCAEwoT\xspace}
\newcommand{\cnnae}{BCAE\xspace}
\newcommand{\ds}{D_\textrm{s}}
\newcommand{\dr}{D_\textrm{r}}
\newcommand{\ls}{\mathcal{L}_\textrm{s}}
\newcommand{\lr}{\mathcal{L}_\textrm{r}}
\newcommand{\rb}{ResBlock\xspace}
\newcommand{\blank}{\hl{\texttt{Blank}}\xspace}

\newcounter{daggerfootnote}
\newcommand*{\daggerfootnote}[1]{%
    \setcounter{daggerfootnote}{\value{footnote}}%
    \renewcommand*{\thefootnote}{\fnsymbol{footnote}}%
    \footnote[2]{#1}%
    \setcounter{footnote}{\value{daggerfootnote}}%
    \renewcommand*{\thefootnote}{\arabic{footnote}}%
    }
\definecolor{color1}{HTML}{AFFFFF}
\definecolor{color2}{HTML}{74DBEF}
\definecolor{color3}{HTML}{5E88FC}
\definecolor{color4}{HTML}{264E86}
\definecolor{color5}{HTML}{FF9A00}

\tikzstyle{mybox} = [
	draw=color1, 
	fill=color1, 
	very thick,
    rectangle, 
    rounded corners, 
    inner xsep=5pt, 
    inner ysep=5pt,
    align=center,
]
\tikzstyle{fancytitle} = [
	fill=color4,
	draw=color4,
	text=white,
	inner sep=2pt,
	rounded corners=3pt, 
]

\newif\ifblind
\blindtrue  
\blindfalse

\title{
Efficient Data Compression for 3D Sparse TPC via Bicephalous Convolutional Autoencoder\\
}

\author{
\ifblind
    anonymous authors
\else
    \IEEEauthorblockN{%
        Yi Huang\IEEEauthorrefmark{1},
        Yihui Ren\IEEEauthorrefmark{1},
        Shinjae Yoo\IEEEauthorrefmark{1}, and
        Jin Huang\IEEEauthorrefmark{2}%
    }%
    \IEEEauthorblockA{\IEEEauthorrefmark{1} \textit{Computational Science Initiative, Brookhaven National Laboratory}, yhuang2, yren, sjyoo@bnl.gov}%
    \IEEEauthorblockA{\IEEEauthorrefmark{2} \textit{Physics Department, Brookhaven National Laboratory}, jhuang@bnl.gov}%
    
\fi
}

\date{}

\begin{document}
\allowdisplaybreaks
\maketitle
\begin{abstract}
    Real-time data collection and analysis in large experimental facilities 
    present a great challenge across multiple domains, including high energy
    physics, nuclear physics, and cosmology. To address this, machine learning (ML)-based
    methods for real-time data compression have drawn significant attention.
    However, unlike natural image data, such as CIFAR and ImageNet that are
    relatively small-sized and continuous, scientific data
    often come in as three-dimensional ($\mathbf{3}$D) data volumes at high rates with high sparsity (many
    zeros) and non-Gaussian value distribution.  
    This makes direct application of popular ML compression 
    methods, as well as conventional data
    compression methods, suboptimal. To address these obstacles, this work introduces a dual-head
    autoencoder to resolve sparsity and regression simultaneously, called \textit{Bicephalous Convolutional AutoEncoder} (BCAE). This method
    shows advantages both in compression fidelity and ratio
    compared to traditional data compression methods, such as MGARD, SZ,
    and ZFP. To achieve similar fidelity, the best performer among the traditional methods
    can reach only half the compression ratio of BCAE. Moreover, a
    thorough ablation study of the BCAE method shows that a dedicated 
    segmentation decoder improves the
    reconstruction. 
\end{abstract}




\section{Introduction}
\label{sec:introduction}

\ifblind
\else

The super Pioneering High Energy Nuclear Interaction eXperiment (sPHENIX) at the Relativistic Heavy Ion Collider (RHIC) will perform high-precision measurements of jets and heavy-flavor observables for a wide selection of high energy nuclear collision systems, elucidating the microscopic nature of strongly interacting matter, ranging from nucleons to strongly coupled quark-gluon plasma~\cite{PHENIX:2015siv}. As illustrated in Figure~\ref{fig:tpc_intro}, a next-generation continuous readout time projection chamber (TPC) is the main tracking detector for sPHENIX. The overall sPHENIX experimental data rate is dominated by the TPC, which reads out at a rate of $O(1)$ Tbps and records at $O(100)$ Gbps to persistent storage~\cite{sPHENIX_TDR}. Therefore, the experiment's storage efficiency and throughput capability can directly benefit from high-fidelity compression algorithms optimized for the TPC data stream. 

\fi

An autoencoder (AE) network is an effective method for data compression~\cite{hinton_reducing_2006,wang_generalized_2014}. 
Compared to traditional data reduction methods, 
such as principal component analysis~\cite{frs_liii_1901} and Isomap~\cite{tenenbaum_global_2000}, 
AE often results in a higher compression ratio and better reconstruction accuracy~\cite{fournier_empirical_2019}. 
Conceptually, AE consists of two neural networks (NNs): 
an encoder and a decoder. 
The encoder maps a high-dimensional data block into low-dimensional compressed representations, 
while a decoder recreates the input data from the code. 
For image data with shift-invariance properties, 
a convolutional autoencoder (CAE) often is used. 
CAE uses a convolutional network~\cite{lecun_backpropagation_1989,krizhevsky_imagenet_2012,simonyan_very_2014,he_deep_2016} 
as the encoder and a deconvolutional network~\cite{zeiler_deconvolutional_2010,zeiler_adaptive_2011} as the decoder. 
Data compression is achieved by having a smaller bit length compared to input.
Unlike other generative NN models, such as a variational autoencoder~\cite{kingma_auto-encoding_2013,razavi_generating_2019,vahdat_nvae_2020} or generative adversarial network (GAN)~\cite{goodfellow_generative_2014,brock_large_2018},
AE emphasizes faithful recreation of the original data point 
instead of the diversity of generated data. 
CAE is not without issues. Reconstructed images often are blurry and sometimes accompanied by checkerboard artifacts~\cite{odena_deconvolution_2016}, which can be especially devastating for sparse data with sharp boundaries.

\ifblind
\else

\begin{figure*}[ht]
    \centering
    \tikzsetnextfilename{tpc_intro}
    \input{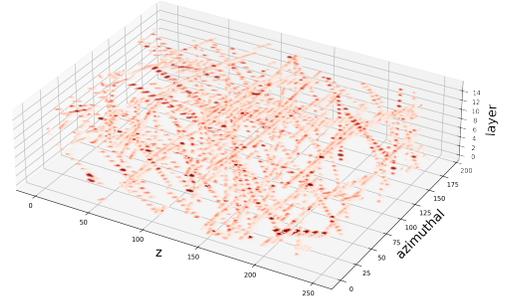}
    \caption{\textbf{sPHENIX experiment with the TPC in yellow, simulated data with particle trajectories in TPC highlighted in red, and $13$~$\mu$s of TPC data presented in a $3$D data frame.}}
    \label{fig:tpc_intro}
\end{figure*}

\fi

\section{Method}
\subsection{Data Gathering and Preparation}
\label{subsec:DataGatheringAndPrep}
\ifblind
\else

We use simulated data for $200$~GeV Au$+$Au collisions detected by the sPHENIX TPC, which is generated based on the HIJING event generator~\cite{Wang:1991hta} and Geant4 Monte Carlo detector simulation package~\cite{Allison:2016lfl} integrated with the sPHENIX software framework~\cite{sPHENIX_Software}. 
sPHENIX's TPC continuously detects thousands of charged particles produced at high-energy Au$+$Au at RHIC, which collide at a rate of $O(100)$~kHz. The ionization charge produced by these particles in the TPC gas volume is drifted, amplified, and collected by $160$k mini pads~\cite{Azmoun:2018ail} and digitized continuously using analog-digital converters (ADCs) inside of the SAMPA v5 application-specific integrated circuit at a rate of $20$~MHz~\cite{Hernandez:2019pqz,sPHENIX_TDR}. As the ionization charge drifts along the $z$ axis at $\sim 8$~cm$/\mu$s, the ADC time series can be translated to the $z$-location dependent ionization charge density. 
All ADC values are $10$-bit unsigned integers ($\in[0, 1023]$), representing the charge density of the initial ionization. The trajectory location is interpolated between the location of neighboring pads using the ADC amplitude. Hence, in a lossy compression, it is important to preserve the relative ADC ratio between the pads. Before readout of the TPC data, the ADC values are zero-suppressed in the SAMPA chips. For this study, we assume a simple zero suppression as ADC$>64$ and set all ADC bins below this threshold to zero. The data streams from SAMPA chips are read out via $960$ $6$-Gbps optical fibers through the FELIX interfaces~\cite{Chen:2019owc} to a fleet of commodity computing servers. The computing capability in the TPC readout path allows a compression algorithm to be applied in the field-programmable gate array in the FELIX interface or the computing servers.
\fi
The detector's TPC mini-pad array is composed of $48$ cylindrical layers that are grouped into the inner, middle, and outer layers, each with $16$ layers.
Detectors in each layer can be unwrapped into a rectangular grid with rows in the $z$ direction and columns in the azimuthal direction.
All layer groups have the same number of rows of detector pads but different numbers of columns. 
Each full three-dimensional ($\mathbf{3}$D) data volume of the outer layer group is in the shape of $(2304, 498, 16)$ in the azimuthal, horizontal, and radial orders. 
To match the segmentation of the TPC readout data concentrator, 
we divide a full frame 
into $12$ non-overlapping sections along the azimuth and halve 
the horizontal direction, resulting in the input data shape of $(192, 249, 16)$.

\subsection{Bicephalous Convolutional AutoEncoder (BCAE)}
\label{subsec:CNNAE}
\begin{figure*}[ht]
    \centering
    \tikzsetnextfilename{cnnae_architecture}
    \def\scale{.85}


\begin{tikzpicture}
\def\summaryScale{.8}
\node[inner sep=0, anchor=north] (summary) at (0, 0) {\input{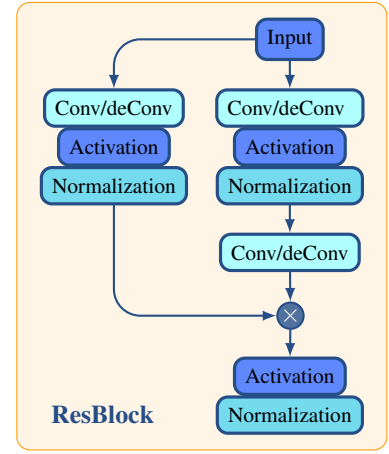}};

\def\scale{.81}
\def\top{1}
\node[inner sep=0, anchor=north west] (block) at ([xshift=.15in,yshift=-.2in]summary.north east) {


\begin{tikzpicture}[scale=\scale, every node/.style={scale=\scale}]
\node[mybox, draw=color4, fill=color3, text opacity=\top] (input) at (0, 0) {Input};
\node[mybox, draw=color4, fill=color1, anchor=north, text opacity=\top] (conv_1) at ([yshift=-.18in]input.south) {
    Conv/deConv 
};
\node[mybox, draw=color4, fill=color3, anchor=north, text opacity=\top] (act_1) at (conv_1.south) {Activation};
\node[mybox, draw=color4, fill=color2, anchor=north, text opacity=\top] (norm_1) at (act_1.south) {Normalization};

\node[mybox, draw=color4, fill=color1, anchor=north, text opacity=\top] (conv_2) at ([yshift=-.18in]norm_1.south) {
    Conv/deConv
};
\node[mybox, draw=color4, fill=color3, anchor=north, text opacity=\top] (act_2) at ([yshift=-.55in]conv_2.south) {Activation};
\node[mybox, draw=color4, fill=color2, anchor=north, text opacity=\top] (norm_2) at (act_2.south) {Normalization};

\node[mybox, draw=color4, fill=color1, anchor=north east, text opacity=\top] (conv_s) at ([xshift=-.2in]conv_1.north west) {
    Conv/deConv
};
\node[mybox, draw=color4, fill=color3, anchor=north, text opacity=\top] (act_s) at (conv_s.south) {Activation};
\node[mybox, draw=color4, fill=color2, anchor=north, text opacity=\top] (norm_s) at (act_s.south) {Normalization};

\draw[->, >=latex, color4, thick] (input) -- (conv_1);
\draw[->, >=latex, color4, thick] (norm_1) -- (conv_2);
\draw[->, >=latex, color4, thick, rounded corners=10pt] (input) -| (conv_s);
\node[circle, draw=color4, fill=color4!80, thick, inner sep=0pt, text=white] (plus) at ([xshift=-.06in]$(conv_2.south)!.4!(act_2.north)$) {\large $\times$};
\draw[->, >=latex, color4, thick] (conv_2) -- (plus);
\draw[->, >=latex, color4, thick, rounded corners=10pt] (norm_s) |- (plus);
\draw[->, >=latex, color4, thick] (plus) -- (act_2);

\node[inner sep=5pt, anchor=south west, color4, text opacity=\top] (label) at (norm_s.west |- norm_2.south) {\large\textbf{ResBlock}};

\begin{scope}[on background layer]
    \def\bw{.15in} 
    \def\bh{.1in} 
    \draw[draw=color5, fill=color5!10, rounded corners=5pt] ([xshift=-\bw, yshift=\bh]input.north -| norm_s.west) rectangle ([xshift=\bw, yshift=-\bh]norm_2.south east);
\end{scope}

\end{tikzpicture}

};
\node[inner sep=0, anchor=south west] (summary_title) at ([yshift=.0in]summary.north west) {{\Large a.} \cnnae architecture summary};
\node[inner sep=0, anchor=south west] (block_title) at (summary_title.south west -| block.north west) {{\Large b.} \cnnae ResBlock};
\end{tikzpicture}

    \caption{
        \textbf{Summary of the \cnnae architecture and the \rb.} 
        Our \cnnae is composed of one encoder and two decoders. 
        The decoders $\ds$ is designated to solve 
        the segmentation problem. 
        The combined output from $\ds$ and $\dr$ is used to solve the regression problem.
    }
    \label{fig:cnnae_architecture}
\end{figure*}
\begin{figure}[ht]
    \centering
    \tikzsetnextfilename{transform_motivation}



\begin{tikzpicture}
\pgfplotstableread[col sep=comma]{figures/data/transform_motivation_hist_original.csv} \histO 
\pgfplotstableread[col sep=comma]{figures/data/transform_motivation_line_original.csv} \lineO 
\pgfplotstableread[col sep=comma]{figures/data/transform_motivation_hist_transformed.csv} \histT 
\pgfplotstableread[col sep=comma]{figures/data/transform_motivation_line_transformed.csv} \lineT 
\pgfplotstableread[col sep=comma]{figures/data/transform_motivation_hist_original_log-log.csv} \histL
\def\barWidth{.016in}
\def\opacity{.7}
\def\color{black}

\begin{groupplot}[
    axis line style={draw=black!75},
	group style={
		group size=2 by 1,
		vertical sep=40pt,
		horizontal sep=23pt,
	},
	width=1.95in,
  	height=1.88in,
	grid,
	grid style={draw=black!15},
  	enlarge x limits={abs=.05in},
  	enlarge y limits={abs=.05in},
  	xtick style={draw=none},
  	xticklabel style={inner sep=2pt},
]
    \nextgroupplot[
        ylabel={density}, 
        xlabel={ADC value},
        ylabel style={yshift=-4pt},
        title style={anchor=south west, at={(-.02, .93)}},
        title={{\Large a.} Original},
        xtick distance=250,
        yticklabel style={
            /pgf/number format/fixed,
            /pgf/number format/precision=2
        },
        scaled y ticks=false,
    ]
        \addplot[
            ybar,
            bar width=\barWidth, 
            draw=\color!50, 
            fill=\color!30, 
            opacity=\opacity,
        ] table[x=x, y=y] {\histO};
        \addplot[thick,draw=black] table[x=x, y=y] {\lineO};
        
        \pgfkeys{/tikz/savenumber/.code 2 args={\global\edef#1{#2}}}
        \coordinate (insetSW) at (rel axis cs:.32,.35); 
        \coordinate (insetNE) at (rel axis cs:.96,.96); 
        \path
            let
                \p1 = (insetSW),
                \p2 = (insetNE),
                \n1 = {(\x2 - \x1)},
                \n2 = {(\y2 - \y1)}
            in [savenumber={\insetwidth}{\n1}, savenumber={\insetheight}{\n2}];

    \nextgroupplot[
        xlabel={$\mathcal{T}\paren{\textrm{ADC value}}$},
        title style={anchor=south west, at={(-.02, .945)}},
        title={{\Large b.} Transformed},
        xtick distance=1,
        yticklabel style={
            /pgf/number format/fixed,
            /pgf/number format/precision=2
        },
        scaled y ticks=false,
    ]
        \addplot[
            ybar,
            bar width=\barWidth, 
            draw=\color!50, 
            fill=\color!30, 
            opacity=\opacity,
        ] table[x=x, y=y] {\histT};
        \addplot[thick,draw=black] table[x=x, y=y] {\lineT};
\end{groupplot}

\begin{axis}[
	grid,
	grid style={draw=black!15},
    at={(insetSW)}, 
    anchor=south west, 
    name=inset, 
    width=\insetwidth,
    height=\insetheight, 
    scale only axis,
    xtick style={draw=none},
    ytick style={draw=none},
    ymode=log,
    ticklabel style={font=\footnotesize, inner sep=1pt},
    axis background/.style={fill=black!5, fill opacity=1},
    ylabel={count},
    xlabel={log ADC value},
    ylabel style={font=\footnotesize, inner sep=1pt},
    xlabel style={font=\footnotesize, inner sep=1pt},
    xtick={0, 2, 4, 6, 8, 10},
    xticklabels={0, 2, 4, 6, 8, 10},
    ymin=1e1,
]
    \addplot[
        ybar,
        bar width=.3*\barWidth, 
        draw=\color!80, 
        fill=\color!50, 
        opacity=\opacity,
    ] table[x=x, y=y] {\histL};
    \coordinate (L) at (axis cs:.2,1e3);
    \coordinate (R) at (axis cs:6,1e3);
    \draw[latex-latex] (L) -- (R);
    \node[inner sep=1pt,fill=black!5,font=\footnotesize] at ($(L)!.5!(R)$) {gap};
\end{axis} 

\end{tikzpicture}

    \caption{
        \textbf{Motivation for input transformation.}
        Panel a: histogram of nonzero ADC values;
        Panel b: histogram of $\mathcal{T}(\textrm{nonzero ADC values})$. 
        The sharp edge around $64$ and long tail are resolved to an extent
        with the transform.
        The inset of Panel a shows the histogram in logarithmic scale of $\log_2(\cdot)$ 
        ADC values, revealing the sparsity of nonzero values and
        a gap between $0$ and $64$.
    }
    \vspace{-.15in}
    \label{fig:transform_motivation}
\end{figure}
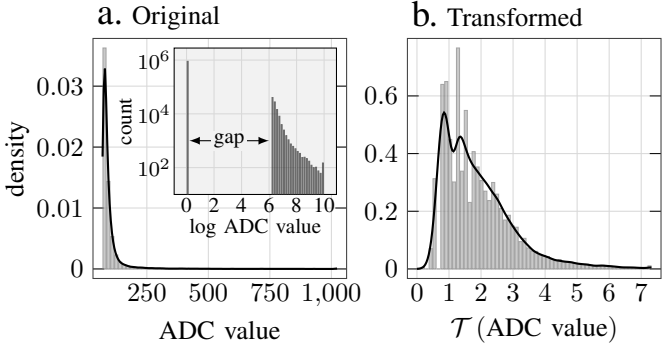

The long tail and sharp edge in the distribution of 
ADC values present a great challenge.
Figure~\ref{fig:transform_motivation}a inset shows the histogram in
logarithmic scale of ADC values, 
revealing the sparsity and a gap between $0$ and $64$.
NNs have difficulties with such distributions~\cite{alanazi2020simulation,hashemi2019lhc},
which we confirm in the ablation studies (Sec.~\ref{subsec:AblationStudy}).
We propose two techniques to solve the problem:
1) reforming the voxel-level regression
into a segmentation (voxel-level binary classification) and 
a regression and 
2) predicting a \textit{transform} of the input data.

\vspace{.03in}
\noindent\textbf{Bicephalous decoder:}
Figure~\ref{fig:cnnae_architecture}a shows that
our \cnnae has two decoders, 
$\ds$ for voxel segmentation and $\dr$ for regression.
From the compressed data, 
the segmentation decoder $\ds$ classifies each voxel 
into two categories: zero-valued or positive-valued. 
The output from $\ds$ is evaluated with the focal loss $\ls$, 
which is designed to deal with an unbalanced data set~\cite{lin2017focal}.
The output from the regression decoder $\dr$ is combined 
with that from $\ds$ and evaluated 
with a mean squared error (MSE) loss $\lr$.

\vspace{.03in}
\noindent\textbf{Predicting a transformed input:}
To deal with the sharp edge in the input distribution 
induced by the zero suppression, 
$\dr$ does not predict an input ADC value, but rather its transform.
The transform has the form: $\mathcal{T}(x) = \log(x - 64) / 6$, 
and let $x$ be a voxel with nonzero value $v_{x}$, 
$\dr$ approximates $\mathcal{T}(v_x)$. 
To motivate the transform,
we plot the histogram of raw and transformed nonzero ADC values in Figure~\ref{fig:transform_motivation}a and b 
(for all ADC values, see the inset of a).
It is evident that the transformed distribution better resembles a normal distribution.
We borrow the idea of input transform 
from~\cite{alanazi2020simulation}, where the technique is applied 
to an electron-proton scattering events simulation
with GAN. 
As in \cite{alanazi2020simulation}, we do not transform the input before feeding it to the network. We assume the network's output (more precisely,
that of $\dr$) is an approximation for the transformed input. 
Before passing the prediction to the loss function, 
we apply the inverse transform $\mathcal{T}^{-1}$.

\vspace{.03in}
\noindent\textbf{Segmentation loss $\ls$: } 
Let $v_{x}$ be voxel $x$'s ADC value, and
the \emph{log ADC value} $v'_x$ is defined to be $\log_2\paren{v_x + 1}$.
Because a nonzero ADC value is $\geq 64$, the true class label of 
$x$ should be $0$ if $v'_x < 6$ or $1$ if otherwise. 
However, to make the loss function differentiable, 
we use the Sigmoid step function
\begin{linenomath}
\[
    \mathcal{S}(v; \mu, \alpha) = \paren{1 + \exp\paren{-\alpha(v - \mu)}}^{-1}
\]
\end{linenomath}
to achieve a ``soft classification'' instead. 
The Sigmoid step function maps $v <\mu$ to a value close to $0$ and $v > \mu$, $1$. 
For this study, we set $\mu = 6$ and $\alpha=20$.
Let $\hat{l}_x$ be the output of $\ds$ for voxel $x$, and
the \textit{focal loss} is defined to be
\begin{linenomath}
\begin{align}
    \lr\paren{\set{\left.\hat{l}_x\right|x};\gamma} =& \frac{1}{M}\sum_{x}-l_x\log_2\paren{\hat{l}_x}\paren{1 - \hat{l}_x}^\gamma \nonumber \\ 
    &-\paren{1 - l_x}\log_2\paren{1 - \hat{l}_x}\paren{\hat{l}_x}^\gamma, \label{eq:clfLoss}
\end{align}
\end{linenomath}
where $M$ is the total number of voxels,
$l_x$ is the soft label, and $\gamma$ is the focusing parameter.
Focal loss is used because we have only $10\%$
of nonzero ADC values on average,
and focal loss is shown to perform well
when positive examples are relatively sparse.
In this study, we set $\gamma=2$.

\vspace{.1in}
\noindent\textbf{Regression loss $\lr$:}
Denoted by $\hat{v}_{x}$ the prediction for voxel $x$ given by $\dr$,
the combined prediction $\tilde{v}_x$ is defined as
\begin{linenomath}
\begin{equation}
    \label{eq:CombinedOutput}
    \tilde{v}_x = 
        \left\{
            \begin{array}{ll}
                \mathcal{T}^{-1}\paren{\hat{v}_x}, & \hat{l}_x \geq h,\\
                0, & \hat{l}_x < h,
            \end{array}
        \right.
\end{equation}
\end{linenomath}
where $h$ is a threshold and the regression loss $\lr$ is defined as
\begin{linenomath}
\begin{equation}
    \lr\paren{\set{\left.\tilde{v}_x, v_x\right| x}; h} = \frac{1}{M}\sum_{x}\paren{\tilde{v}_x - v_x}^2,
\end{equation}
\end{linenomath}
where $M$ is the total number of voxels.

\vspace{.1in}
\noindent\textbf{Architectural details:}
\cnnae{s} are constructed 
using the \rb in Figure~\ref{fig:cnnae_architecture}b. 
The \rb design is inspired by the 
residual block proposed in~\cite{he2016deep}.
However, unlike a residual block in~\cite{he2016deep} 
that preserves the input size, 
we downsample and upsample with \rb. 
On the main path of a \rb, 
we have (de)convolution, normalization, activation, 
and an additional (de)convolution layer.
We downsample or upsample at the first (de)convolution layer and use
kernel size $3$ and stride $1$ uniformly 
for the second.
We also only increase or decrease the number of channels 
at the first (de)convolution layer on the main path 
while maintaining it at the second (de)convolution layer.
The three layers on the \rb sidetrack are exactly the same
as the first group of layers on the main path, and the encoder has four \rb{s}.
The encoder has four \rb{s} and ends with a $1\times1\times1$ convolution layer with $8$ channels,
while  $\ds$ and $\dr$ have three \rb{s} plus a deconvolution layer as shown in Figure~\ref{fig:cnnae_architecture}a. 
$\ds$ also has a Sigmoid layer to constrain output values between $0$ and $1$.
For activation and normalization, we use leaky rectified linear activation function ReLU with negative slope $0.1$ and 
instance normalization.

\vspace{.03in}
\noindent\textbf{Training}:
We implement the \cnnae with \texttt{PyTorch}, 
using $960$ sections (see Sec.~\ref{subsec:DataGatheringAndPrep}) for training and $320$ for testing.
We train the \cnnae with batch size $32$ for $2000$ epochs. 
For the optimizer, we use the AdamW 
with $\beta_1, \beta_2 = 0.9, 0.999$ and $0.01$ weight decay. 
We set the initial learning rate to be $0.01$ and decrease it by $5\%$ every $20$ epochs. 
To match the magnitudes of $\ls$ and $\lr$,
we use the following approach to combine the losses: let $\rho^{t}_\textrm{s}$ and $\rho^{t}_\textrm{r}$ be 
the segmentation and regression losses of epoch $t$. Then, the loss function at epoch $t+1$ is defined as 
$\mathcal{L}^{t+1} = \paren{\rho^{t}_\textrm{r} / \rho^{t}_\textrm{s}}\ls + \lr$.

\section{Results}
\label{sec:Results}


\subsection{\cnnae Compression Ratio}
\label{subsec:CompressionRatio}
The ADC values originally are $10$-bit integers saved as $16$-bit unsigned integers. 
\cnnae runs with $32$-bit floats, and the compressed data output from the encoder as $32$-bit floats. 
To increase the compression ratio, we save 
the compressed data as $16$-bit floats. 
The decoders then can either run with half-precision ($16$-bit floats) 
or upcast the compression data to $32$-bit floats before running.
As such, we can calculate \cnnae's compression ratio 
using only the input and output shapes. 
The input shape is $(1, 192, 249, 16)$ (See Sec.~\ref{subsec:DataGatheringAndPrep}). 
The output shape is $(8, 13, 17, 16)$. 
Hence, the compression ratio of \cnnae equals $\paren{1 \times 192 \times 249 \times 16} / \paren{8\times 13 \times 17 \times 16} = 27.04$.

\subsection{Comparison with Three Existing Compression Algorithms}
\label{subsec:ComparisonWithExistingAlgs}

\begin{figure}[ht]
    \centering
    \tikzsetnextfilename{comparison_mse_scatter}


\begin{tikzpicture}

\pgfplotstableread[col sep=comma]{figures/data/comparison_mse_scatter.csv} \data

\begin{groupplot}[
    axis line style={black!60},
	group style={
		group size=1 by 1,
		vertical sep=40pt,
		horizontal sep=30pt,
	},
	width=2.8in,
  	height=2.2in,
	grid,
  	enlarge x limits={abs=.1in},
  	enlarge y limits={abs=.1in},
  	xtick style={draw=none},
]
    \nextgroupplot[
        legend cell align=left,
        legend style={
            draw=black!20, 
            fill=white, 
            fill opacity=.6, 
            text opacity=1,
            anchor=north west, 
            at={(.05, .94)},
        },
        xlabel={compression ratio},
        ylabel={MSE},
    ]
        \addplot[only marks, mark=*, mark size=5pt, draw=color3!80!black, fill=white, thick, forget plot] coordinates {(27.0539, 696.8245)};
        \addplot[only marks, mark=triangle*, mark size=6pt, draw=color4!80!black, fill=white, thick, forget plot] coordinates {(23.9551, 373.5443)};
        \addplot[only marks, mark=diamond*, mark size=6pt, draw=color2!80!black, fill=white, thick, forget plot] coordinates {(19.2562, 330.0340)};
        \addplot[only marks, mark=square*, mark size=5pt, draw=color5!80!black, fill=white, thick, forget plot] coordinates {(27, 219)};
        
        \addplot[only marks, mark=*, mark size=2.5pt, draw=color3!60!black, fill=color3] table[x=MGARD_x, y=MGARD_y] {\data};
        \addplot[only marks, mark=triangle*, mark size=3pt, draw=color4!60!black, fill=color4] table[x=SZ_x, y=SZ_y] {\data};
        \addplot[only marks, mark=diamond*, mark size=3pt, draw=color2!60!black, fill=color2] table[x=ZFP_x, y=ZFP_y] {\data};
        \addplot[only marks, mark=square*, mark size=2.5pt, draw=color5!60!black, fill=color5] coordinates {(27, 219)};
        \legend{MGARD, SZ, ZFP, \cnnae}
        
        \draw[-stealth, thick, red] (axis cs: 27, 50) --node[pos=.5, below, black, fill=white, fill opacity=.4, text opacity=1] {\small higher better} (axis cs: 34.8, 50);
        \draw[-stealth, thick, red] (axis cs: 35.5, 275) --node[pos=.5, below, rotate=90, black, fill=white, fill opacity=.4, text opacity=1, ] {\small lower better} (axis cs: 35.5, 75);
\end{groupplot}
\end{tikzpicture}

    \caption{
        \textbf{Relation between MSE and compression ratio.} 
    }
    \label{fig:comparison_mse_scatter}
\end{figure}
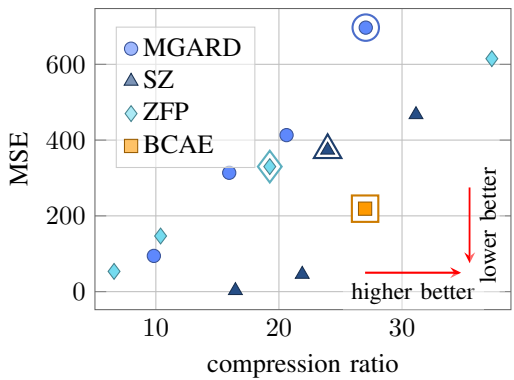

\begin{figure*}[t]
    \centering
    \tikzsetnextfilename{comparison_histogram}
    \def\scale{.9}
    \input{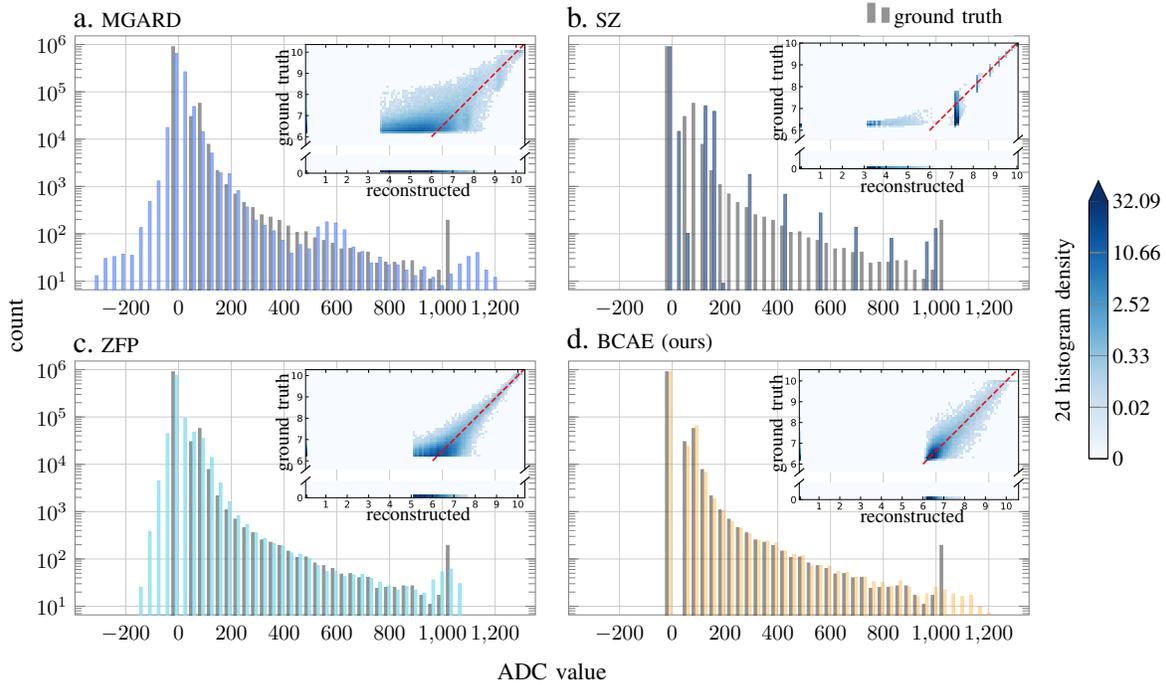}
    \caption{
        \textbf{Histogram of reconstructed ADC values for the four compression methods.} 
        To compare with \cnnae at compression ratio $27$, 
        we fix MGARD at ratio $27.05$, 
        SZ at ratio $23.96$, 
        and ZFP at ratio $19.26$. 
        The reconstructed distribution of \cnnae (Panel d) 
        is closer to that of the ground truth than the other three methods at similar or smaller compression ratios. 
        Both MGARD and ZFP have reconstructed values below zero, 
        while SZ's approximation is overly simplified and distorted. 
        Of note, a spike on the high end (ADC$=1023$) 
        is due to the $10$-bit limitation of the ADC chip, 
        which outputs an overflow value with the max ADC number. 
        \cnnae decompressed value may exceed this instrumental limit.
    }
    \label{fig:comparison_histogram}
\end{figure*}

MGARD~\cite{chen2020accelerating}, or MultiGrid Adaptive Reduction of Data, 
is a technique for multilevel lossy compression of scientific data based 
on the theory of multigrid methods~\cite{ainsworth2018multilevel}. 
MGARD takes an $\ell_\infty$ error bound as the parameter, 
and the compression is successful 
if the maximum absolute difference between the raw 
and reconstructed values is below the bound.
SZ~\cite{di2016fast,tao2017significantly,liang2018error} 
is an error-controlled lossy compression algorithm 
optimized for high-compression ratios. 
SZ provides five error-bounding modes, including 
absolute and relative errors and their combinations.
ZFP~\cite{lindstrom2014fixed} is a compressed format for representing two- ($2$) to four ($4$)-dimensional 
arrays that exhibit spatial correlation.
ZFP provides three lossy compression modes: fixed-rate, fixed-precision, and fixed-accuracy. 
The fixed-accuracy mode is the same as the $\ell_\infty$ error-bound mode.
ZFP requires padding to make dimensions a multiple of $4$.
The compressed data of all three benchmarking algorithms are saved in binary, 
and their compression ratios are calculated 
as the averaged ratio of input size (as $16$-bit floats) to output file size. 
For comparison, we use the $\ell_\infty$ 
error-bound mode, which is provided by all three algorithms. 
Because the $\ell_\infty$ error bound does not 
offer direct control over compression ratio,
we determine the relation between MSE and compression ratio
by surveying a range of $\ell_\infty$ bounds for each algorithm.  
Specifically, for each $\ell_\infty$ bound, 
we run the algorithm on $100$ sampled $3$D sections then calculate the averaged compression ratios and MSEs. 


To compare reconstruction distributions 
with \cnnae at compression ratio $27$, 
we fix MGARD at ratio $27.05$, 
SZ at ratio $23.96$, and ZFP at ratio $19.26$.
Figure~\ref{fig:comparison_mse_scatter} summarizes the results and shows that \cnnae achieves a much lower MSE
than benchmarking algorithms at similar compression ratios. 
We also demonstrate one ($1$)-dimensional and $2$D histograms in Figure~\ref{fig:comparison_histogram}. 
For $1$D histograms, we plot the ground truth distribution of ADC values in gray.
Both MGARD and ZFP have reconstructed values below zero, 
while SZ's approximation is overly simplified and distorted. 
Because of the gap in the ground truth distribution between $0$ and $64$, 
we can post-process the reconstructed values of MGARD, SZ, and ZFP by thresholding. 
Explicitly, we find a threshold $h$ and map each 
reconstructed value $<h$ to $0$ while keeping those $>h$ unchanged. 
The threshold values, 
$11$ for MGARD, $8$ for SZ, and $32$ for ZFP, 
are chosen to minimize the MSEs.
We plot $2$D histograms after the post-processing on log ADC values.
We omit the portions of $2$D histograms that represent the gap in the ground truth.
We calculate the MSE, mean absolute error of log ADC value (log MAE), and Peak signal-to-noise ratio (PSNR)
after the post-processing and record the results in Table~\ref{tab:comparison}. 
All evaluations are done on raw ADC values.

\begin{table}[ht]
    \centering
    \caption{Performance comparison}
    \begin{tabular}{lrrrr}
        \toprule
        {} &    \thead{Compr.~ratio}$\uparrow$ & \thead{MSE}$\downarrow$ & \thead{log MAE}$\,\downarrow$ & PSNR$\uparrow$\\
        \midrule
        MGARD   & $\mathbf{27}$  & $626.28$  & $1.213$   & $3.223$ \\
        SZ      & $24$  & $369.69$  & $0.302$   & $3.452$ \\
        ZFP     & $19$  & $219.48$  & $0.267$   & $3.678$ \\
        \midrule
        \single & $27$ & $227.61$ & $0.349$ & $3.703$ \\
        \double & $27$ & $230.59$ & $0.193$ & $3.706$ \\
        \cnnae  & $\mathbf{27}$  & $\mathbf{218.44}$  & $\mathbf{0.185}$   & $\mathbf{3.724}$ \\
        \bottomrule
    \end{tabular}
    \label{tab:comparison}
\end{table}

\subsection{Ablation Study}
\label{subsec:AblationStudy}

\begin{figure*}[t]
    \centering
    \tikzsetnextfilename{ablation}
    \def\scale{1}
    \input{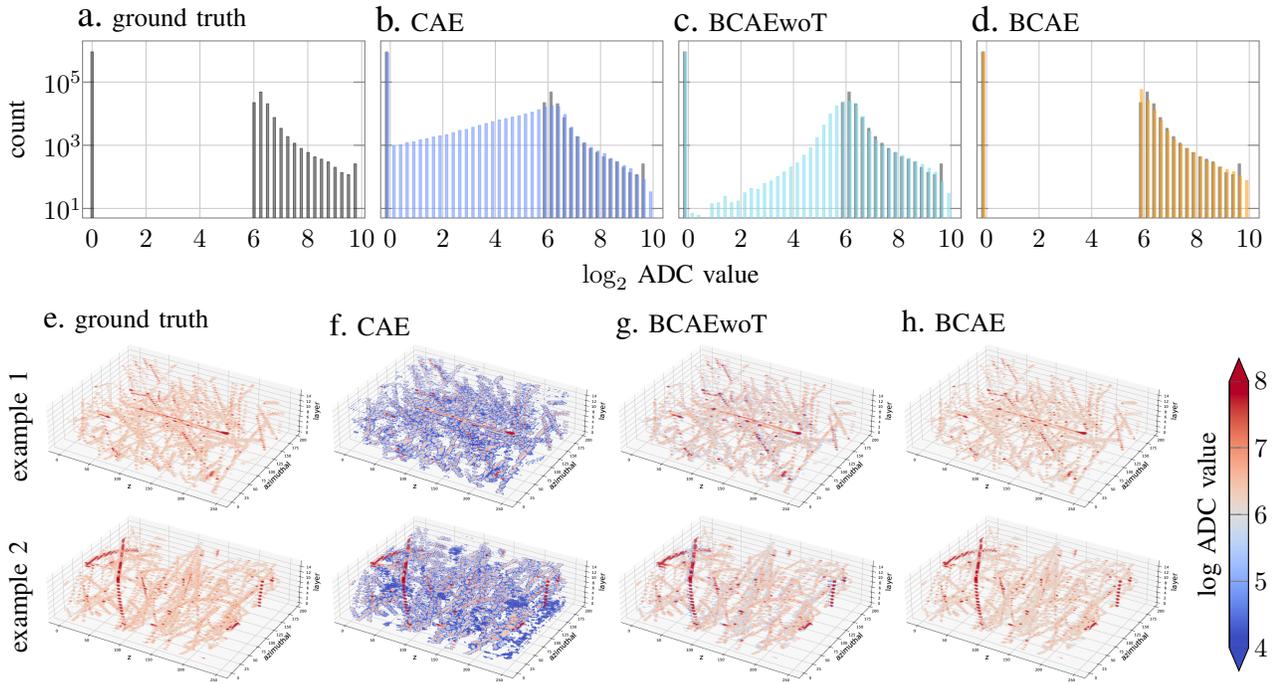}
    \caption{
        \textbf{Ablation study.}
        We compare the \cnnae to regression-only CAE and BCAE without transform (\double). The reconstructed distribution of \cnnae (Panel d) 
        most closely resembles the ground truth distribution, 
        and the reconstructed \cnnae images (Panel h) also have the highest quality.
    }
    \label{fig:ablation}
\end{figure*}

We compare the \cnnae to regression-only CAE and BCAE without transform (\double).
\double~is obtained by adding a ReLU layer 
to $\dr$ of \cnnae, and \single is gleaned by removing $\ds$ from \double.
The combined prediction $\tilde{v}_x$ \double determines for $x$ is
\begin{linenomath}
\begin{equation}
    \label{eq:CombinedOutputWithoutTransform}
    \tilde{v}_x = 
    \left\{
        \begin{array}{ll}
            \hat{v}_x, & \hat{l}_x \geq h,\\
            0, & \hat{l}_x < h.
        \end{array}
    \right.
\end{equation}
\end{linenomath}
In Figure~\ref{fig:ablation}a-d, we show the histograms of true log ADC values $v'_x$ and 
reconstructed ones by \single, \double, and \cnnae of 1 million sampled voxels.
Figure~\ref{fig:ablation}b shows that \single fills in the gap, meaning it significantly blends and blurs around the segmentation, 
which is confirmed by the $3$D visualization in Figure~\ref{fig:ablation}f. 
\double approximates the input distribution better 
because a fraction of voxels with small 
ADC values are classified as zero.
In Figure~\ref{fig:ablation}g, we see that \double offers 
more clear-cut reconstructed images.
Figure~\ref{fig:ablation}d shows that the reconstructed distribution 
of \cnnae most closely resembles the original distribution,
while reconstructed \cnnae images (Figure~\ref{fig:ablation}h) also have the highest quality. 

Because $\ds$ outputs a continuous value in $[0, 1]$, 
we may adjust $h$ in Eq.~\eqref{eq:CombinedOutput} and \eqref{eq:CombinedOutputWithoutTransform} 
after training to decrease MSE further. 
Notably, with $h=.4$ for \double and $h=.46$ for \cnnae, 
we get the lowest respective MSEs on training data.
Test results in Table~\ref{tab:comparison} are obtained by applying the new thresholds
instead of the $.5$ used in training.
Furthermore, as the downstream application of TPC hit position determination 
focuses the ratios of ADC values between neighboring voxels,
we also care about MAE in logarithmic scale of ADC values, where  
\cnnae achieves the lowest score. 

\section*{Conclusion}
We have developed BCAE, a deep learning method for sparse ($\mathbf{3}$D) data compression. 
Compared to conventional compression methods, BCAE achieves higher fidelity and higher compression ratio.
Yet, as with all deep learning methods, a trained BCAE model is data set specific and requires unsupervised training. 
We also demonstrate the effectiveness of our dual-head decoder design by conducting an ablation study that shows both logarithmic transform and dedicated segmentation decoder are important to match the ADC value distribution in logarithmic scale, which is critical in our application. 

In the future, we want to integrate a differentiable proxy of sPHENIX downstream analysis criteria as part of the loss function to enable end-to-end training. In this way, the compression method becomes downstream task-aware and can better serve the overall scientific need. 
The segmentation decoder $\ds$ also can be trained with domain information to identify noise voxels. 
We expect \cnnae can become a powerful tool that may sensitively preserve the signal voxel while filtering out the noise voxels in the raw data,
which will improve the compression ratio and perform data reduction in a single set of algorithms. 

\section*{Acknowledgment}
The authors are grateful for the significant assistance received from
Charity Plata in the editing of this paper.
This work was supported in part by the Office of Nuclear Physics within the U.S. DOE Office of Science under Contract No. DESC0012704, and Brookhaven National Laboratory under Laboratory Directed Research \& Development No. 19-028.



\bibliographystyle{plain}
\bibliography{bibliography,AutoEncoder,MultiTaskLearn}

\end{document}


\allowdisplaybreaks
\maketitle
\section*{Appendix}
The encoder $E$ has four \rb{s} with 
kernel sizes $\mathbf{f}^{E}_i$, 
strides $\mathbf{s}^{E}_i$, 
and number of input/output channels $C^{E}_{\textrm{in/out}}$
listed as follows: 
\begin{linenomath}
\begin{subequations}
\label{eq:cnnae_encoder}
\begin{align}
    &\mathbf{f}^{E}_1 = (4, 5, 3),\, C^{E}_{\textrm{in}, 1} = 1,\,\,\, C^{E}_{\textrm{out}, 1} = 8,  \label{eq:cnnae_encoder1}\\
    &\mathbf{f}^{E}_2 = (3, 3, 3),\, C^{E}_{\textrm{in}, 2} = 8,\,\,\, C^{E}_{\textrm{out}, 2} = 16, \label{eq:cnnae_encoder2}\\
    &\mathbf{f}^{E}_3 = (3, 3, 3),\, C^{E}_{\textrm{in}, 3} = 16,\, C^{E}_{\textrm{out}, 3} = 32, \label{eq:cnnae_encoder3}\\
    &\mathbf{f}^{E}_4 = (3, 4, 3),\, C^{E}_{\textrm{in}, 4} = 32,\, C^{E}_{\textrm{out}, 4} = 32, \label{eq:cnnae_encoder4}\\
    &\mathbf{s}^{E}_i = (2, 2, 1), i = 1, 2, 3, 4.
\end{align}
\end{subequations}
\end{linenomath}
Both decoders, $\ds$ and $\dr$, have three \rb{s} with
kernel sizes $\mathbf{f}^{D}_i$, 
strides $\mathbf{s}^{D}_i$, 
and number of input/output channels $C^{D}_{\textrm{in/out}}$
listed as follows: 
\begin{linenomath}
\begin{subequations}
\label{eq:cnnae_decoder}
\begin{align}
    &\mathbf{f}^{D}_{1} = (3, 4, 3),\, C^{D}_{\textrm{in}, 1} = 8,\, C^{D}_{\textrm{out}, 1} = 8, \label{eq:cnnae_decoder1}\\
    &\mathbf{f}^{D}_{2} = (3, 3, 3),\, C^{D}_{\textrm{in}, 2} = 8,\, C^{D}_{\textrm{out}, 2} = 4, \label{eq:cnnae_decoder2}\\
    &\mathbf{f}^{D}_{3} = (3, 3, 3),\, C^{D}_{\textrm{in}, 3} = 4,\, C^{D}_{\textrm{out}, 3} = 2, \label{eq:cnnae_decoder3}\\
    &\mathbf{f}^{D}_{4} = (4, 5, 3),\, C^{D}_{\textrm{in}, 4} = 2,\, C^{D}_{\textrm{out}, 4} = 1, \label{eq:cnnae_decoder4}\\
    &\mathbf{s}^{D}_i = (2, 2, 1), i = 1, 2, 3, 4.
\end{align}
\end{subequations}
\end{linenomath}
All output paddings are $0$ for the deconvolutions.